# Feature Selection with Annealing for Forecasting Financial Time Series

Hakan Pabuccu, Adrian Barbu

## Abstract

Stock market and cryptocurrency forecasting is very important to investors as they aspire to achieve even the slightest improvement to their buy-or-hold strategies so that they may increase profitability. However, obtaining accurate and reliable predictions is challenging, noting that accuracy does not equate to reliability, especially when financial time-series forecasting is applied owing to its complex and chaotic tendencies. To mitigate this complexity, this study provides a comprehensive method for forecasting financial time series based on tactical input–output feature mapping techniques using machine learning (ML) models. During the prediction process, selecting the relevant indicators is vital to obtaining the desired results. In the financial field, limited attention has been paid to this problem with ML solutions. We investigate the use of feature selection with annealing (FSA) for the first time in this field, and we apply the least absolute shrinkage and selection operator (Lasso) method to select the features from more than 1,000 candidates obtained from 26 technical classifiers with different periods and lags. Boruta (BOR) feature selection, a wrapper method, is used as a baseline for comparison. Logistic regression (LR), extreme gradient boosting (XGBoost), and long short-term memory (LSTM) are then applied to the selected features for forecasting purposes using 10 different financial datasets containing cryptocurrencies and stocks. The dependent variables consisted of daily logarithmic returns and trends. The mean-squared error for regression, area under the receiver operating characteristic curve, and classification accuracy were used to evaluate model performance, and the statistical significance of the forecasting results was tested using paired *t*-tests. Experiments indicate that the FSA algorithm increased the performance of ML models, regardless of problem type. The FSA hybrid models showed better performance and outperformed the other BOR models on seven of the 10 datasets for regression and classification. FSA-based models also outperformed Lasso-based models on six of the 10 datasets for regression and four of the 10 datasets for classification. None of the hybrid BOR models outperformed the hybrid FSA models. Lasso-based models, excluding the LR type, were comparable to the best models for six of the 10 datasets for classification. Detailed experimental analysis indicates that the proposed methodology can forecast returns and their movements efficiently and accurately, providing the field with a useful tool for investors.

## 1 Introduction

The financial stock market was introduced in the 1790s, and since then, many investors have gained significant profits via their prediction capabilities and optimized portfolios. Stock market investors have generally used the latest technology to increase their forecasting accuracy (Yoo et al., 2005), and their enthusiasm and ambition for more profit have led to new and powerful tools for this purpose. After the financial crisis of 2008 and its effects on investors, institutions, academia, and governments, cryptocurrencies have received attention as a viable alternative source of investment revenue, despite their highly speculative behaviors. Recently, especially during the COVID-19 pandemic, cryptocurrency and stock investments have received great scrutiny from investors and researchers worldwide. Since then, large increases in trading volumes have increased the importance of rapid forecasting capabilities.

Both stock markets and cryptocurrencies depend on numerous macroeconomic factors, such as political changes, economic outlooks, investor expectations, and global events. Financial time-series data are regarded as noisy and ambiguous to many, as they have nonlinear dynamic structures and chaotic movements. Although predictions can be made using nonlinear equations and multidimensional operators, the computing power needed to provide rapid capability is elusive. This capability gap has long-supported the optimistic efficient market hypothesis (EMH; Malkiel & Fama, 1970), leading to continued attempts to create innovative and effective forecasting algorithms. According to the EMH, all stock market indices can be effectively defined, and future forecasts can be made using extant trading datasets, even in real time. Hence, efficient market strategies can be developed across the investor-to-corporation spectrum via rigorous scientific analyses (Leung et al., 2000). The greater the diligence, the lower the potential market risk. Although this makes intuitive sense, even when including a wide variety of external factors, pragmatists have long realized the need to consider the stochastic aspects of market performance. Many of the smartest investors hedge their margin risk based on an assumed chaos factor, which is a tacit way of admitting defeat in the face of exponential analytical complexity. According to

Alves et al. (2020), the dynamic aspect of market efficiency remains too poorly understood to provide time-dependent high-accuracy predictions. This phenomenon has been borne out with the advent of cryptocurrency trading markets (Alves et al., 2020). Sigaki et al. (2019) reported that cryptocurrencies are informationally efficient within a tunable time window and can theoretically fit the EMH model. This has been disputed (Yoo et al., 2005), and has increased the urgency of breakthrough algorithmic models.

In classical regression analyses, univariate and multivariate financial time series were used on stock index datasets that were preprocessed to eliminate perceived chaotic behaviors and contain uncertainties, basically eliminating nonlinear relationships. Although this process allowed regression analysis, the information loss applied to the data resulted in poor results. In addition to finding new ways to condition datasets, computer scientists began investigating the application of neurological theory to computational methods. Machine Learning (ML) theory has existed for decades. For example, artificial neural network (ANN) theory was formulated by Hebb (1949) and the computational perceptron was invented by Rosenblatt (1958). However, the hardware and computing capacity has only recently allowed functional applications (McCulloch & Pitts, 1990); Minsky & Papert, 1969). A period of rapid growth began with the reinvention of the back-propagation algorithm by Rumelhart et al. (1986), who made the ANN a prominent actor in producing stochastic predictability by allowing nonrestrictive assumptions to be made on the data. Hassan et al. (2007), Kara et al. (2011), Kimoto et al. (1990), Olson and Mossman (2003), Wei (2016) have since produced fantastic ANN breakthroughs, and Hsu et al. (2009), Huang et al. (2005), and Kumar and Thenmozhi (2005) have yielded breakthrough results with secure vector machines.

The nonsequential data type used in multilayer perceptron (MLP) models and the temporal dependent structure of the input data used with recurrent neural networks (RNNs) are quite a bit more convenient than MLPs when applied to financial time-series forecasting. Kamijo and Tanigawa (1990) applied an RNN model using Tokyo Stock Exchange data and obtained a 93.8% prediction accuracy. The vanishing gradient problem specific to deep neural networks was addressed by the introduction of Long-Short Term Memory (LSTM) networks. Cho et al. (2014) and Di Persio and Honchar (2017) applied a LSTM model with a dropout function to Google stock prices, and similar applications were invented by other researchers (Bao et al., 2017; Zhao et al., 2017; Pawar et al., 2019).

The financial market capitalization of US-listed domestic companies was more than USD 40T in 2020, which is nearly two times the US gross domestic product (World Bank 2023).

This paper proposes a two-step forecasting mechanism for financial data in which a feature-selection method is applied to select a subset of relevant features, and a highly complex model is applied to forecast returns and movements. The Boruta (BOR) feature-selection method (Ghosh et al., 2022), which works well with big data via a wrapper model, and feature selection with annealing (FSA; Barbu et al., 2017) and Lasso (Tibshirani, 1996) algorithms, as embedded methods, are applied for feature selection. The purpose of this construction is to improve forecasting performance. The main contributions of this paper are as follows:

- We formalized the financial time-series forecasting problem as one of the regressions and classifications to forecast return and return movements, respectively.
- To determine the most effective features and forecast returns and return movements, we provide a hybrid forecasting approach based on feature selection and ML techniques.
- Empirical results on 10 financial datasets demonstrate that FSA hybrid models outperform other ML models when forecasting prices and price movements.
- The FSA algorithm improves the forecasting performance of the selected ML algorithms.
- The Lasso and BOR algorithms yield comparable results to FSA for some cases.
- These results present evidence of the validity of the EMH for the stock and cryptocurrency markets.

The remainder of this paper is organized as follows. Section 2 presents a literature review, and Section 3 explains the datasets used and the preprocessing techniques applied. Section 4 introduces the feature selection and forecasting algorithms used in the experiments. Section 5 presents the experimental results and performance comparisons of different feature selection and forecasting models. Finally, Section 6 presents a discussion, and Section 7 concludes with conclusions and future work.

**2 Literature Review**

*2.1 Financial time-series forecasting*

Shah et al. (2019) organized the financial time-series prediction techniques into four categories: statistical methods, ML, pattern recognition, and sentiment analysis. Financial time-series forecasting problems can be categorized into classification and regression types. In this section, statistical methods and ML models are applied within the scope of this research. Statistical models simplify assumptions to obtain theoretical guarantees, resulting in simple structures and potentially lower performance than ML techniques. ML models are more generalized, make fewer simplifying assumptions, and offer better model-fitting capabilities (Fang et al., 2021). According to Patel et al. (2015a) and Henrique et al. (2019), the two most widely used techniques are ANN and support vector regression (SVR), which are supervised methods; unsupervised methods are less popular. To forecast the future value of financial assets and find reasons for asset behaviors, numerous ML techniques have been employed, such as SVM (Akyildirim et al., 2021), SVR (Kara et al., 2011), random forest (RF ;Patel et al., 2015a), and the convolutional neural network (CNN; Tsantekidis et al., 2017). These works show that it is possible to use ML techniques to make more accurate forecasts and as alternative approaches to conventional techniques.

Some leading studies used financial data and statistical models for financial forecasting's, such as FitzPatrick (1932), Beaver (1966), Altman (1968), Ohlson (1980). These studies were based on bankruptcy prediction using financial ratios as predictors. FitzPatrick (1932) provided an early business failure prediction model, identifying five steps to review the problem. Beaver (1966) recognized that financial ratios do not have the same effect on the possibility of failure. Altman (1968) used linear discriminant analysis (LDA) to improve model performance, reaching an accuracy level of 94% for the first year, 72% for the second, and 48% for the third. After applying conventional statistical techniques, neural networks and other ML algorithms provide new financial forecasting directions. ANNs, SVMs, and decision trees, for example, can learn the trends of stock or cryptocurrency returns from historical data. As such, they provide significant analytical data. Related works presented as regression and classification problems follow.

*2.2 Regression-based forecasting*

Bernal et al. (2012) applied an Echo State Network RNN to forecast S&P 500 stock prices using technical indicators, such as moving averages. Their technique outperformed the Kalman filter as a benchmark based on testing error statistics from 50 different financial datasets. Roondiwala et al. (2017) applied an LSTM to forecast Nifty prices by using open, high, low, and close values as input variables. The root mean-squared error was used as the performance metric. Lahmiri and Bekiros (2019) discussed a financial problem as a regression using an LSTM and a generalized regression neural network (GRNN) using daily Bitcoin, Digital Cash, and Ripple prices were as input features. Lahmiri and Bekiros (2019) provided an LSTM with significantly better performance than the GRNN. Han et al. (2020) used a nonlinear autoregressive model with an exogenous variable (NARX) technique to predict USD Bitcoin returns on a daily frequency. Bitcoin returns were also used as an input feature, and the authors reported that the NARX model could predict trends, but not breaks.

*2.3 Classification-based forecasting*

Ballings et al. (2015) compared the performance of ANN, Logistic Regression (LR), SVM, and k-nearest neighbors with AdaBoost and RF models. They reported the RF algorithm was the best option to forecast stock price movements based on the area under the curve (AUC) and cross-validation metrics. Di Persio and Honchar (2017) reached a level of accuracy of 72% using Google stock prices by comparing different RNN model performances, including an LSTM, a gated recurrent unit, and a basic RNN. Nousi et al. (2019) applied several ML techniques, such as a single hidden layer feed-forward neural network, MLP, autoencoder, and bag-of-feature algorithms, to forecast the mid-price movements of stock prices, achieving good success. Smuts (2019) applied an LSTM using an input set containing Telegram chat groups focusing on Bitcoin, Google trends, price volumes, and Bitcoin–Ethereum trades. The results showed that Telegram data were a good predictor for Bitcoin. However, the weekly Google trend data were better for Ethereum. Borges and Neves (2020) examined the most traded volume of 100 cryptocurrency prices from Binance with a 1-min frequency using returns and technical indicators as input. The authors regarded this as a classification problem and applied LR, RF, SVM, gradient tree boosting, and an ensemble of the mentioned algorithms. Chen et al. (2020) used 5-min and daily Bitcoin price indices and trading prices in USD to measure LR, LDA, RF, ANN, LSTM, and extreme gradient-boosting (XGBoost) algorithm performance. Four blockchain information variables, trading variables,

Google trend search volume indices, and gold spot prices were used as input features. The authors reported that the LSTM model had the best accuracy level of 67% for 5-min data, and LR and LDA had the best accuracy levels of 65% for daily data. Sun et al. (2020) used 42 cryptocurrencies' daily data and applied a light gradient-boosting machine (GBM), SVM, and RF to forecast the movement of prices. Sun et al. (2020) reported that light GBM outperformed other ML techniques based on accuracy. Fang et al. (2021) reported that the mid-price movement of Bitcoin could be predictable at an accuracy level of 78% using an LSTM. Fister et al. (2021) presented a new approach to finding a superior strategy for daily trading on a portfolio of stocks by comparing traditional trading strategies to a set of LSTM networks. This research showed that the LSTM significantly outperforms traditional trading strategies regardless of whether they are based on universal or specific features as inputs. Like many return movement forecasting studies, the LSTM network is a significant alternative to forecasting optimal stock portfolios. Basher and Sadorsky (2022) looked at forecasting Bitcoin price movements to predict possible trends in a specific period and to plan asset allocations. Their research investigated the importance of cryptocurrency forecasting using different input variables, such as inflation rate, interest rate, and market volatility. The authors applied tree-based ML techniques to forecast the direction of Bitcoin prices and reported accuracy levels between 75 and 80%.

McNally et al. (2018) used Bitcoin price open, close, high, and low values and the hash and difficulty rates of the blockchain for different periods to construct an input set. The problem was considered a regression and classification type, and LSTM and Bayesian RNNs were used as learning algorithms. The authors reported that 20 days comprised the best period for the RNN and 100 days for the LSTM. Atsalakis et al. (2019) proposed a hybrid neuro-fuzzy controller (PATSOS) model to forecast Bitcoin, Ethereum, Ripple, and Litecoin prices, considering the problem to be one of classification and regression. PATSOS was observed to have significantly better forecasting results than the other benchmark models selected in their study. Moreover, PATSOS predicted returns significantly better for the buy/hold strategies based on the predicted signs of the selected technical indicators.

### *2.4 Feature selections for forecasting*

Feature selection is an important problem for financial time-series forecasting, especially in the social sciences. An effective feature-selection process can improve the generalizability of forecasting models by removing irrelevant features from the input space. However, this problem has not received much attention in the financial literature. According to Niu et al. (2020), there are three types of feature-selection methods: embedded types with restricted models (e.g., linear), which are not accurate enough; filter-based methods that are not powerful; and wrapper methods that are too time-consuming for many input variables. Each method has advantages and limitations and can be applied to financial time-series forecasting problems.

Tyralis and Papacharalampous (2017) proposed a feature-selection model based on an RF algorithm to suggest an optimal set of input variables. The authors used two different time-series datasets and compared the results to standard benchmarks. As a result, the RF model used for feature selection showed better performance when using a few short-lag input features. Valente and Maldonado (2020) proposed an approach for time-series forecasting by adapting the SVR for feature selection. The authors used an autoregressive integrated moving average model with feature selection to forecast energy loads. A total of 1,700 lags were used for high-frequency data, and fewer than 400 were used for daily data. The proposed methodology showed a slightly better performance than the selected conventional techniques. Niu et al. (2020) proposed a two-stage, multiobjective wrapper-based feature selection to determine the optimal input space for deep learning models. The mean absolute error, mean-squared error (MSE), and mean average percentage error statistics were used to evaluate the model performance. The authors reported that the proposed feature-selection model significantly improved forecasting performance. Ghosh et al. (2022) proposed an ensemble feature-selection algorithm to forecast stock prices using national stock exchange data in India based on the COVID-19 effects. Spot prices, market sentiments, sectoral outlooks, crude price volatilities, and exchange rate volatility features were used in their study. The authors proposed a structural model to determine the effects of selected variables comprising BOR and regularized RF algorithms. Regularized greedy forest and deep neural network algorithms were used with principal component analysis and autoencoders. It was shown that the importance of input features depends on the particular stock and the period under consideration.

When conducting an overall evaluation, some general comments should be made. If the model structure is well defined, it becomes possible to obtain accurate forecasts. When comparing the ML algorithms for classification and regression, it is clear that neural network models, especially LSTMs, commonly provide superior forecasts. Owing to its mathematically complex structure and its design for handling vanishing gradient problems, the researchers expected that the LSTM algorithm would outperform MLP and RNN networks. However, there were no theoretical guarantees of obtaining the best forecasts in all cases due to the existence of influencing factor accuracy. Another algorithm that provides accurate forecasts in the literature is the RF model, which is a tree-based construct that obtained accuracy levels of around 75% in some studies. Other relatively new models, such as autoencoders, attention mechanisms, and natural language processing (NLP) algorithms, which have not received significant attention thus far, may provide notable results for classification and regression problems. The reviewed studies showed that the ML techniques provided good results with lower computational costs while adapting easily to provide superior estimations.

Forecasting performance can differ based on the dataset used and the preprocessing techniques used to prevent overfitting. Financial time-series studies generally use the same dataset with different time periods, namely "open," "close," "high," and "low" values at first. Then, the variable construction process using these values can differ from the others, affecting forecasting performance. A limited number of studies have considered the volatility of the dataset used, providing more comprehensive contributions to the relevant literature by increasing forecasting accuracy. Models using two or more forecasting algorithms provide better predictions, but require more user proficiency and experience. The more important point is choosing and setting the input and output variable mapping procedures. Wrong mapping procedures can lead to higher accuracy or vice versa, but lower realism. To the best of our knowledge, a limited number of studies based on feature selection exist in financial fields, despite providing useful results and decreasing the input feature space dimension, numerical calculations, and calculation time. However, several studies have been performed using a limited number of features without feature-selection processes based on mathematical or statistical models. These studies used the most common feature set as the input. From our experience, the variable selection process is a very important tool for improving model performance owing to slight improvements in forecasting performance, leading to the avoidance of economic losses. In addition to feature selection, the feature-generation process may help improve forecasting performance by using deep learning methods.

The main contribution of this paper is that it illuminates the importance of feature selection for ML forecasting problems in both classification and regression settings. Our process uses 10 financial time-series datasets that include cryptocurrency and stock market information. This study investigates the predictability of selected stock and cryptocurrency returns using ML techniques with and without feature selection as regression problems for binary buy and sell strategies. The study evaluates several ML techniques, namely LRs, ANNs, CNNs, LSTM, and XGBoost, after applying FSA (Barbu et al., 2017), Lasso (Tibshirani 1996), and BOR (Ghosh et al., 2022) as benchmarks.

## 3 Data Preparation

In this research, three information sources were selected:

- Historic prices from Yahoo Finance using the Python Finance API—The collected data contained the date, opening price, high price, low price, closing price, adjusted closing price, and volume information from a specific date interval, depending on availability. A different number of observations was obtained for each dataset, owing to data availability and trading hours. Summary information about the dataset is listed in Table 1.

- Technical indicators—Four classes of technical indicators were used, including momentum, volume, volatility, and trend. Trend indicators show the direction of market movement and are oscillators as they cycle between high and low values. Momentum indicators measure model strength and forecast possible reverse movements. Volume measurements show how volume changes over a specific period. Volatility indicators measure the extent to which value changes occur over a specific period. Market volatility, momentum, and trend algorithms are designed to maximize profits (Kumar & Patil, 2018; Salkar et al., 2021). Technical analysis provides investors and analysts with comprehensive summaries and forecasts by calculating technical indicators. However, most analysts use a very limited number of indicators when providing comments and recommendations. Furthermore, different periods and lags can assist with forecasting. A total of 26 technical indicators were gathered using the Python Technical Analysis Library (TA-Lib). Some indicators have subindicators, such as Bolinger Bands, which have three. When calculating technical indicators, six different periods were used: 2, 4, 8, 16, 32, and

64 for each convenient indicator, and one for some entries. A total of 185 features were obtained from these methods.

- Lag operator—As reported in (Bação et al., 2018; Hyun et al., 2019; Omane-Adjepong & Alagidede, 2019; Sebastião & Godinho 2021), the stock and cryptocurrency returns are highly interconnected at different frequencies and lags. Thus, the first five lag periods (1, 2, 3, 4, and 5) were used for all 185 indicators and added to the feature pool. A total of 925 features were obtained using this operator. As a result of these calculations, five features were removed from the datasets because they contained too many NAN values, resulting in 1,105 features.

Using these three sources, a feature pool was constructed from historical stock and cryptocurrency data to apply feature-selection algorithms and remove irrelevant features from the model. Data preprocessing was applied at each analytical step to obtain clean datasets. This involved removing some rows (samples) and features from datasets and some infinite values coming from various calculations. These processes resulted in the loss of data points; however, algorithmic performance was not significantly affected due to the otherwise sufficient data. Hence, different numbers of observations were obtained for each dataset. There were also differences between the stock and cryptocurrency data samples regarding the hours of operation and availability. The input datasets were normalized between zero and one for feature selection and forecasting, and 1,105 features were obtained for the input feature space, as listed in Table 1.

Table 1: Dataset information

| Data | Code | Start date | End date | Observations | Features |
|---|---|---|---|---|---|
| Tether | TTH | 3/24/2018 | 4/21/2022 | 1477 | 1105 |
| Etherium | ETH | 11/09/2017 | 5/17/2022 | 1513 | 1105 |
| Bitcoin | BTC | 4/21/2017 | 4/21/2022 | 1690 | 1105 |
| Ripple | RPL | 3/24/2018 | 5/31/2022 | 1525 | 1105 |
| Tesla | TSL | 4/21/2017 | 4/21/2022 | 1112 | 1105 |
| Nasdaq | NSDQ | 11/28/2017 | 5/16/2022 | 1123 | 1105 |
| Nikkei225 | NKK | 5/17/2017 | 5/17/2022 | 1022 | 1105 |
| New York Stock Exc. | NYSE | 11/28/2017 | 5/16/2022 | 1085 | 1105 |
| Standard &Poors 500 | S&P500 | 5/17/2017 | 5/16/2022 | 1121 | 1105 |
| Gold | GLD | 11/28/2017 | 5/16/2022 | 1101 | 1105 |

We considered two prediction targets: regression and classification. As such, there were two options for selecting the regression targets as an output: closing prices and logarithmic returns. The most commonly used indicator was preferred as an output to reach more realistic results. For classification, a binary output (Eq. 1) was used to represent the up and down movements of the logarithmic returns. This is a very common process for representing price and return trends, which equate to economic profits or losses. The up and down movements were coded as {−1,1} for FSA and {0, 1} for the forecasting models.

$$y_t = \begin{cases} 1, if\ return_t > return_{t-1} \\ -1, else \end{cases}. \quad (1)$$

We considered the forecasting problem of predicting the target at time *t* from the input features at time *t − 3*. In our experiments, we observed that when trying to predict the target at time *t* from the input features at time *t* or *t − 1*, very high accuracy and high $R^2$ values were obtained for regression, indicating that the research problem was too simple and realistically unsuitable. This is why we settled on time *t* from the features at time *t − 3*, where lower accuracy and higher MSE values were obtained; however, they showed more reliable and realistic results. The partitioning ratio of data into training and testing depends on several factors, such as user experience and the amount of data. In many cases, the testing data proportion fell between 30 and 20%, depending on the number of samples and data type. Thus, there is no precise way to choose a data partitioning rate. In this paper, the data were partitioned into 70% for training and 30% for testing for all models after doing many preexperiments, and the same randomization process was used for all datasets to make the comparative results truly comparable.

**4 Problem and Methodology**

Many ML financial time-series forecasting articles focused on improving classification accuracy or decreasing errors for regression problems, but they largely ignored feature selection, which is important for generalizability. Unlike traditional statistical methods, an excess of features and dependent variables theoretically improves ML performance, given that hardware and other resources are provided. In addition to feature selectin, another overlooked problem relates to financial time-series forecasting due to high dimensionality and data noise (Längkvist et al., 2014).

In ML circles, a naïve prediction accuracy near 100% is highly scrutinized, as it will likely lead to inaccurate predictions. This is caused by the highly influential nature of training bias and, often, by the user's lack of experience. To formally combat this in this work, we take extra measures to insure highly accurate input–output mapping, which is key to eliminating bias. As a result, we applied a two-step procedure comprising feature selection and forecasting, each using an ML algorithm. This reflects the basic hybrid methodology used in this paper, as discussed below.

*4.1 Feature selection*

Feature selection involves identifying the most relevant features to solve a specific problem when building an ML model. Doing so helps improve model generalizability while reducing computational complexity and the time costs associated with obtaining training data. For the current task, we applied the FSA method (Barbu et al., 2017), which is a highly efficient feature-selection model that provides statistically true feature recovery and convergence. Moreover, it is easy to implement and handle nonlinearities to a certain extent. The FSA problem is formulated as a constrained optimization problem, as shown in Eq. 2:

$$\beta = argmin_{|\{j, \beta_j \neq 0\}| \leq k} L(\beta), \qquad (2)$$

where $k$ is the number of relevant features, and the loss function, $L(\beta)$, is differentiable with respect to $\beta$. The key idea of the algorithmic design includes using an annealing plan to lessen greediness when reducing the dimensionality from $M$ (initial number of features) to $k$ (desired number of features) and gradually removing the most irrelevant features to facilitate computation. The algorithm is usually set with $\beta = 0$ and follows two steps. The first includes parameter updates using gradient descent, and the second removes some features based on coefficient values.

While running the annealing plan, the coefficient vector dimension was reduced until $|\{j, \beta_j \neq 0\}| \leq k$. The FSA parameters (i.e., learning rate, annealing parameter, and number of epoch*s*) are presented in Table 2. The MSE and logistic loss functions were used for regression and classification estimates, respectively.

Table 2: FSA parameter values

| Parameters | Levels |
|---|---|
| Learning rate $\eta$ | 0.01-0.1 |

| Annealing parameter $\mu$ | 0,1,20,50,100,300,500,1000 |
|---|---|
| Number of epoch | 50,100,200,300,500 |

We also evaluated the Lasso algorithm for feature-selection purposes. Lasso was proposed by Tibshirani (1996) for parameter estimation and model selection in regression analysis tasks and is a type of penalized least square estimation method that uses the L1 penalty function. Lasso is defined as follows in Eq. 3:

$$\hat{\beta}^{lasso} = arg\min_{\beta}\left\{\frac{1}{2}\sum_{i=1}^{N}\left(y_i - \beta_0 - \sum_{j=1}^{p}x_{ij}\beta_j\right)^2 + \lambda\sum_{j=1}^{p}|\beta_j|\right\}. \tag{3}$$

The idea with Lasso is that the $L_1$ penalty sets many coefficients to zero; hence, it is a very useful tool for feature-selection purposes. Apart from FSA and Lasso, we also evaluated a BOR feature-selection process (Ghosh et al., 2022) as the benchmark model. BOR uses an RF with a feature importance measure to select the relevant features from the input space (Kursa & Rudnicki, 2010). We used the existing scikit-learn implementation of the BOR algorithm available at https://github.com/scikit-learn-contrib/boruta_py.

Noting that FSA and Lasso are linear selection methods used to select features for nonlinear models, the BOR method is based on a nonlinear model. The argument for using a linear model for feature selection is that financial data are very noisy and that a nonlinear model might be too flexible. That is, it may learn irrelevant patterns from noise, resulting in poor feature selection.

### 4.2 Forecasting

This section provides an overview of the ML methods used to forecast the financial time series in this paper (i.e., LR, ANN, CNN, LSTM, and XGBoost). Previous research has sought to determine the parameter levels needed for the most effective practices. Hence, the best parameter combination was found for each model separately. It is quite difficult to guess which parameter set is best for the data used. Hence, while selecting the best parameters, predetermined evaluation metrics were considered. The levels of the parameters for each method are presented in the following sections.

### 4.2.1 LR

LR is a popular technique to model discrete probabilities (i.e., binary or multinomial) and outcomes indicated by the class label. LR is a multiple regression model with a dependent variable, $y\epsilon 0.1$, for binary classification, where $y \epsilon 0,1,\ldots,n$ for multinomial classification. The dependent variable is predicted from the input feature vector, $x = (x_1, \ldots, x_p) \epsilon R^p$. The LR model can be expressed as follows (Eq. 4):

$$\log \frac{P(y = 1|x)}{1-P(y = 1|x)} = \beta_0 + \beta_1 x_1 + \ldots + \beta_p x_p. \tag{4}$$

Parameters $\beta = \beta_0, \ldots, \beta_p$ are learned via the optimization of the negative log-likelihood loss function, which can also have an $L_1$ or $L_2$ penalty term to improve generalization. We used the $c$ parameter, which represents the inverse of regularization strength, where a smaller $c$ specifies stronger regularization. The Python Scikit-Learn 1.1.1 library was used for LR implementation. The parameter set levels (e.g., penalty and solver) used in the study are given in Table 3.

Table 3: LR parameter levels

| Parameters | Levels |
|---|---|
| Penalty | 0.01-0.1 |
| Inverse regularization strength parameter $c$ | 0.1, 0.3, 0.5, 1 |
| Solver | lbfgs, liblinear |
| Tolerance | 0.0001 |

### 4.2.2 ANN

ANNs are commonly used to distinguish the triggers of stock or cryptocurrency price changes. An MLP's flexible structure has been successful in many studies (Kara et al., 2011; Patel et al., 2015b), indicating that MLP might have an advantage over traditional statistical models. A feed-forward neural network model was used in this study with different numbers of hidden layers and neurons.

The log sigmoid is used in the output layer, and a tangent sigmoid with a rectified linear unit function is used in the hidden layer of the activation function. Stochastic gradient descent and ADAM optimization algorithms are used for ANN training, and the loss functions include the MSE loss for regression and the binary cross-entropy loss for classification. A comprehensive set of parameter combinations in the ranges presented in Table 4 was explored to determine the best combinations. The Python Keras 2.9.0 library was used to implement the ANN.

Table 4: ANN parameter levels

| Parameters | Levels |
|---|---|
| Learning rate $\eta$ | 0.01-0.1 |
| Momentum constant (c) | 0.1-0.9 |
| Number of epochs | 50, 100, 200, 500 |
| Number of hidden layers | 1, 2, 3 |
| Number of neurons for each hidden layer | 10, 20, 50, 100 |
| Activation function | ReLU, tanh, sigmoid |
| Optimizer | SGD, ADAM |

*4.2.3 CNN*

CNNs are among the most popular ML techniques, performing well on different regression and classification problems (Qian et al., 2020). Each neuron receives an input signal from the input feature space and operates as an ANN. The CNN structure allows a connection to locally share weights to effectively determine local features (Zhang et al., 2021). One of the main differences between an ANN and a CNN is the field of usage. A CNN is generally used in pattern recognition tasks with 2D or 3D images and 1D sounds. CNNs have convolutional layers, pooling layers, and often fully connected layers with an output layer. However, the main components are the convolutional and pooling components. The convolutional layer extracts local features, and the pooling layer reduces the dimension of the input space to avoid overfitting (Zhang et al., 2021). The inputs to the networks include time-series data features in the form of a 1D signal. The Python Keras 2.9.0 library was used for the implementation of our CNN, whose technical information, parameters, and levels are listed in Table 5.

Table 5: CNN parameter levels

| Parameters | Levels |
|---|---|
| Convolution layers | Conv1D layer |
| Number of hidden layers | 1,2,3 |
| Number of hidden neurons | 20, 30, 40, 50 |
| Kernel size | 2,3,4,5,6 |
| Pooling layer | MaxPooling1D Layer |
| Pool size/Stride | 2/1 |
| Activation function | ReLU, tanh, sigmoid |
| Optimizer | SGD, ADAM |

*4.2.4 LSTM*

The LSTM network developed by Hochreiter and Schmidhuber (1997) is an RNN extension that was redesigned to handle the vanishing gradient problem. An LSTM consists of a memory cell to store the information coming from inputs through the three gates: input gate $i(t)$, forget gate $f(t)$, and output gate $o(t)$. An LSTM can be formulated as follows (Xue et al., 2020):

$$i_t = \sigma(W_{xi}x_t + W_{hi}h_{t-1} + W_{ci}c_{t-1} + b_i), \tag{5}$$

$$f_t = \sigma(W_{xf}x_t + W_{hf}h_{t-1} + W_{cf}c_{t-1} + b_f), \tag{6}$$

$$c_t = f_t c_{t-1} + i_t \tanh(W_{xc}x_t + W_{hc}h_{t-1} + b_c), \tag{7}$$

$$o_t = \sigma(W_{xo}x_t + W_{ho}h_{t-1} + W_{co}c_{t-1} + b_o), \tag{8}$$

$$h_t = o_t \tanh(c_t), \tag{9}$$

where $x_t$ and $h_t$ are the input and hidden state, respectively at time **t**, where $W_j$, $W_i$, and $W_0$ are the weight matrices of the corresponding components. $b_f$, $b_i$, and $b_0$ are the corresponding bias parameters, and $\sigma$ is the activation function.

We adopted LSTM as our ML method for predicting the future based on past information without assuming noise forms. Most importantly, an LSTM can possibly capture nonlinear features beyond the time series. The LSTM hyperparameters include the number of layers, *L*, and the number of training epochs, *E*. Because a linear structure is a special case of a feed-forward neural network, one should expect that the LSTM will perform at least as well as an MLP. The Python Keras 2.9.0 library was used for this implementation. The parameter levels, activation functions, and optimizers are presented in Table 6.

Table 6: LSTM parameters

| Parameters | Levels |
| --- | --- |
| Activation function | ReLU, tanh, sigmoid |
| Optimizer | SGD, ADAM |
| Number of hidden layers | 1,2 |
| Number of hidden neurons | 10, 20, 30, 40, 50 |
| Number of training epoch | 50 |
| Dropout rate | 0.2, 0.3 |

### *4.2.5 XGBoost*

The XGBoost algorithm is a decision tree-based ML algorithm developed by Chen and Guestrin (2016). It is currently used in finance (Nobre & Neves, 2019), energy (Fatahi et al., 2022), and healthcare fields (Wu et al., 2022). Compared with other gradient-boosting algorithms, XGBoost has the following advantages. It effectively considers missing information, prevents overfitting, and has a lower computation time owing to parallel processing capabilities. XGBoost uses a weak learner to minimize its loss function and optimize the objective function (Luo et al., 2021). For any differentiable convex loss functions, $\iota = \mathbb{R} \times \mathbb{R} \to \mathbb{R}$, the objective function, $\sigma(\phi)$, is defined in Eq. 10:

$$\sigma(\phi) = \sum_{i=1}^{n} \iota(\hat{y}_i, y_i) + \sum_k \Omega(f_k), \tag{10}$$

where $\Omega(f_k) = \gamma T + \frac{1}{2}\lambda\|\omega\|^2$ is a regularization term, $y_i$ indicate the labels, $\omega$ is the weight of a leaf, and T is the number of leaves in the tree. The squared loss function for regression and the logistic loss function were used for classification. The learning rate was set to 0.1, and the number of estimators was set to 100, anticipating that the problem could be solved with iterative methods.

### *4.3 Evaluation criteria*

During feature selection, the MSE loss function in Eq. 11 was used for regression, and the binary cross-entropy loss function in Eq. 12 was used for training. If $y_i$ and $p_i$ represent the true and predicted values, respectively, for example *I*, then

$$MSE = \frac{1}{n}\sum_{i=1}^{n}(y_i - p_i)^2, \tag{11}$$

$$CE = -\frac{1}{n}\sum_{i=1}^{n}(y_i \log(p_i) + (1 - y_i)\log(1 - p_i)). \tag{12}$$

To evaluate and compare model performance, the MSE metric was used for the regression, and accuracy and area under the receiver operating characteristic (ROC) curve metrics based on the confusion matrix were used for classification, both based on the set-aside test set.

Accuracy is the most commonly used metric for binary classification. In this case, the accuracy metric is defined by Eqs. 13 and 14, based on the elements of the confusion matrix, which include true positive (TP), false positive (FP), true negative (TN), and false negative (FN) results, as illustrated in Figure 1.

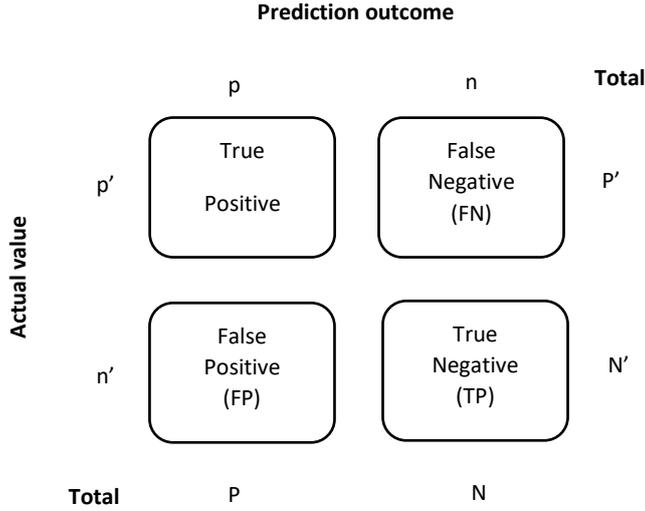

Figure 1. True positives (TPs), false positives (FPs), true negatives (TNs), and false negatives (FNs) of the standard confusion matrix.

$$Accuracy = \frac{TP + TN}{TP + TN + FP + FN}, \qquad (13)$$

$$Recall = \frac{TP}{TP + FN}, \qquad (14)$$

The ROC curve is the curve of the recall versus FP rate for all possible threshold values. The AUC is a scalar between zero and one. Binary classifier performance increases as the AUC approaches one.

## 5 Results

In this section, the experimental results are presented to compare the performance of feature selection and forecasting models. The first phase considered the feature-selection process. The FSA, Lasso, and BOR algorithms were applied to reduce the dimensionality of the input space. BOR was the benchmark model used to evaluate FSA and Lasso performance based on predetermined evaluation criteria. The second phase applied forecasting models on all or selected features to obtain FSA, Lasso, and BOR forecasts of the return movements of selected stocks and cryptocurrencies. LR, ANN, CNN, LSTM, and XGBoost models were used as forecasting models, and paired *t*-tests were performed for statistical significance measurements at the $p < 0.05$ level. In this study, Python v.3.8.13 was used.

### *5.1 Feature-selection results*

### *5.1.1 FSA results*

Because the data were highly dimensional with different numbers of samples for each dataset related to market conditions and data availability, the FSA algorithm (Barbu et al., 2017) with a linear model was applied, and the results are presented in Table 7 alongside the respective parameter combinations. Parameters $\mu$, $s$, and $\eta$ are the most important parameters of the FSA algorithm. The number of relevant features and annealing parameters ($\mu$) generally differs in regression and classification problems. They are usually $\mu = 300$, apart from the Tether, Bitcoin, and Ripple datasets in both regression and NKK for classification. The number of iterations was set to 300 for all models.

Table 7: Results obtained by the linear FSA method (Barbu et al., 2017)

| | Regression | | | | | | Classification | | | | | |
|---|---|---|---|---|---|---|---|---|---|---|---|---|
| Data | $k$ | $\mu$ | $s$ | $\eta$ | $N^{iter}$ | MSE | $k$ | $\mu$ | $s$ | $\eta$ | $N^{iter}$ | Accuracy |
| TTH | 10 | 10 | 0,01 | 0,01-0,01 | 300 | 0.121 | 10 | 1000 | 0,01 | 0,3-0,01 | 300 | 0,501 |
| ETH | 30 | 300 | 0,01 | 0,01-0,01 | 300 | 26.21 | 30 | 300 | 0,01 | 0,01-0,01 | 300 | 0,512 |
| BTC | 30 | 200 | 0,001 | 0,1-0,1 | 300 | 21.42 | 35 | 1000 | 0,01 | 0,01-0,01 | 300 | 0,523 |
| RPL | 30 | 100 | 0,01 | 0,01-0,01 | 300 | 35.73 | 10 | 100 | 0,01 | 0,001-0,001 | 300 | 0,532 |
| TSL | 30 | 300 | 0,01 | 0,005-0,01 | 300 | 14.64 | 30 | 300 | 0,01 | 0,1-0,1 | 300 | 0,507 |
| NSDQ | 40 | 300 | 0,01 | 0,1-0,1 | 300 | 2.468 | 10 | 300 | 0,01 | 0,1-0,1 | 300 | 0,748 |
| NKK | 60 | 300 | 0,01 | 0,01-0,01 | 300 | 1.526 | 60 | 100 | 0,01 | 0,1-0,001 | 300 | 0,519 |
| NYSE | 20 | 300 | 0,01 | 0,1-0,01 | 300 | 1.775 | 10 | 300 | 0,01 | 0,1-0,01 | 300 | 0,499 |
| S&P500 | 25 | 300 | 0,01 | 0,01-0,001 | 300 | 1.711 | 15 | 300 | 0,01 | 0,01-0,01 | 300 | 0,530 |
| GLD | 40 | 300 | 0,01 | 0,01-0,01 | 300 | 1.227 | 45 | 300 | 0,01 | 0,1-0,1 | 300 | 0,538 |

The FSA model reduced the high-dimensional feature space by more than 95%, achieving the same or higher accuracy results on the test set. The accuracy levels of the FSA for each dataset were quite low, around 50%, apart from the NSDQ data. The accuracy level of the NSDQ was 74.8%, which was the best value for classification problems, as shown in Table 7.

*5.1.2 Lasso results*

Lasso is a commonly used powerful feature-selection method in finance. It was applied in this study to reduce the input feature space dimension for an effective forecasting process and to compare the results with other methods. As seen in Tables 7 and 8, the FSA accuracy results were higher, and its MSE values lower than that of Lasso. In addition to these results, Lasso produced a lower MSE but lower accuracy compared with the BOR models. The efficiency of the feature-selection methods was evaluated by comparing the hybrid forecasting results with the ML algorithm-only results.

Table 8: Results obtained using the Lasso feature-selection method (Tibshirani, 1996).

|  | Regression | | Classification | |
| --- | --- | --- | --- | --- |
| Data | Number of relevant feature | MSE | Number of relevant features | Accuracy |
| TTH | 12 | 0.177 | 105 | 0.463 |
| ETH | 16 | 26.72 | 82 | 0.491 |
| BTC | 15 | 17.53 | 83 | 0.461 |
| RPL | 12 | 36.63 | 82 | 0.461 |
| TSL | 12 | 18.26 | 79 | 0.443 |
| NSDQ | 25 | 2.418 | 131 | 0.734 |
| NKK | 12 | 1.516 | 75 | 0.420 |
| NYSE | 11 | 1.800 | 64 | 0.441 |
| S&P500 | 36 | 1.532 | 70 | 0.456 |
| GLD | 32 | 1.168 | 80 | 0.453 |

*5.1.3 BOR results*

The BOR feature-selection method, which is quite popular in the forecasting literature, was evaluated and compared with the FSA and Lasso methods. As seen in Table 9, the regression MSEs of the BOR method were higher than those of the FSA, and their accuracy values were lower than those of the FSA. However, BOR provided better results with higher accuracy than Lasso for classification.

Table 9: Results obtained by the Boruta feature-selection method (Ghosh et al., 2022)

|  | Regression | | Classification | |
| --- | --- | --- | --- | --- |
| Data | Number of relevant feature | MSE | Number of relevant features | Accuracy |
| TTH | 82 | 0.361 | 13 | 0.498 |
| ETH | 15 | 34.464 | 67 | 0.401 |
| BTC | 12 | 24.823 | 88 | 0.467 |
| RPL | 19 | 43.044 | 66 | 0.451 |
| TSL | 14 | 20.803 | 28 | 0.434 |
| NSDQ | 28 | 2.901 | 15 | 0.705 |
| NKK | 6 | 1.905 | 17 | 0.423 |
| NYSE | 13 | 2.416 | 18 | 0.451 |
| S&P500 | 18 | 1.857 | 20 | 0.412 |
| GLD | 14 | 1.406 | 117 | 0.428 |

*5.2 Return forecasting*

Four regression algorithms were tested on the 10 financial time-series datasets, which were used to forecast selected stock returns and cryptocurrency and gold prices. The MSE was used for model comparisons, and the results are presented in Table 10.

Table 10: Return forecasting—MSE values for the 10 datasets evaluated

| Model/Data | TTH | ETH | BTC | RPL | TSL | NSDQ | NKK | NYSE | S&P500 | GLD |
| --- | --- | --- | --- | --- | --- | --- | --- | --- | --- | --- |
| Null | 0.178 | 26.25 | 17.53 | **34.63** | 18.27 | **2.418** | **1.519** | 1.813 | 1.532 | 1.164 |
| ANN | 0.177 | **26.21** | 17.50 | **34.62** | 18.27 | **2.428** | **1.515** | **1.780** | 1.550 | 1.165 |
| CNN | 0.177 | **26.18** | 17.49 | **34.63** | 18.29 | **2.418** | **1.516** | 1.800 | 1.532 | 1.138 |
| LSTM | 0.209 | 27.01 | **17.12*** | 35.99 | 18.44 | **2.435** | 1.713 | 1.944 | 1.673 | 1.206 |
| XGB | 0.756 | 26.98 | 18.16 | 34.92 | 19.22 | 2.463 | 1.644 | 1.877 | 1.540 | 1.174 |
| FSA | **0.121*** | 26.21 | 21.42 | 35.73 | **14.64*** | 2.468 | **1.526** | **1.775** | 1.711 | 1.227 |
| LAS | 0.177 | 26.72 | 17.53 | 36.63 | 18.26 | **2.418** | **1.516** | 1.800 | 1.532 | 1.168 |
| FSA-ANN | 0.177 | **26.17*** | 17.50 | **34.33*** | 18.18 | **2.398*** | **1.509*** | **1.773*** | **1.517*** | 1.167 |
| FSA-CNN | 0.172 | 26.22 | 17.52 | **34.63** | 18.26 | **2.418** | **1.519** | 1.804 | 1.532 | **1.053*** |
| FSA-LSTM | **0.168** | 26.98 | **17.12** | 35.99 | 18.39 | **2.431** | 1.719 | 1.938 | 1.679 | 1.207 |
| FSA-XGB | 0.224 | 26.61 | 17.97 | 35.69 | 18.21 | 2.486 | 1.567 | 1.894 | 1.539 | 1.197 |
| BOR-ANN | 0.177 | 26.68 | 17.71 | 34.98 | 18.29 | **2.417** | **1.520** | 1.821 | 1.691 | 1.165 |
| BOR-CNN | 0.178 | 26.81 | 17.54 | **34.63** | 18.22 | **2.418** | **1.527** | 1.805 | 1.531 | 1.162 |
| BOR-LSTM | 0.213 | 27.71 | 17.89 | 35.99 | 18.39 | **2.432** | 1.692 | 1.970 | 1.680 | 1.199 |
| BOR-XGB | 0.218 | 27.69 | 17.92 | 34.99 | 19.58 | 2.466 | 1.617 | 1.946 | 1.632 | 1.204 |
| LAS-ANN | 0.181 | 26.54 | 26.68 | **34.63** | 18.19 | 2.417 | **1.526** | 1.809 | 1.561 | 1.152 |
| LAS-CNN | 0.265 | 26.39 | 28.53 | 38.76 | 21.05 | **2.423** | 1.866 | 1.906 | 1.533 | 1.163 |
| LAS-LSTM | 0.180 | 26.66 | 28.96 | 35.21 | 18.31 | **2.421** | 1.560 | 1.835 | 1.733 | 1.182 |
| LAS-XGB | 0.212 | 26.96 | 18.07 | 35.65 | 18.57 | 2.506 | 1.601 | 1.926 | 1.555 | 1.189 |

Paired *t*-tests were used to compare the best results from each column (shown in bold with a star) with other results. Results that were not significantly worse ($p > 0.05$) than the best result are also shown in bold, representing the top-performing group.

The null model, which always predicts the average return of the training set, was also evaluated as a baseline comparison. One can easily see that the FSA algorithm improved forecasting performance in most cases. As seen in Table 10, at least one of the FSA models was in the top-performing group (bold) for all nine datasets, excluding the TSL dataset. The FSA linear model provided the best results with the lowest MSE values for the TSL dataset and detected significant differences with the other models. Additionally, the FSA linear models were in the top-performing group in five of the 10 datasets.

If we compare the FSA-based model performances to the other models, we can see that at least one FSA-based model significantly outperformed all non-FSA ML models on three of the 10 datasets (i.e., TTH, S&P500, and GLD). In contrast, the non-FSA models, including Lasso and BOR, never outperformed the FSA models in any dataset.

The null model was in the top-performing group on three datasets (i.e., RPL, NSDQ, and NKK), which means that no model could significantly improve the performance of the null model on these datasets. In the six remaining datasets (i.e., ETH, BTC, RPL, NSDQ, NKK, and NYSE), the FSA-based models were in the top-performing group, alongside some non-FSA models.

When we compared the ML and hybrid FSA results with the hybrid Lasso and BOR results, none of the hybrid Lasso and BOR models outperformed the other ML and hybrid model performances in terms of MSE. However, the Lasso-based hybrid models were in the top-performing group alongside the FSA-based models on three datasets (i.e., RPL, NSDQ, and NKK). The BOR models showed statistically equal performance as the best FSA-based models on three of 10 datasets, as did the Lasso-based models. Moreover, when examining the seven datasets in which the null model was not within the top-performing group, we found that the Lasso and BOR models were not in the top-performing group in any of the seven datasets. Therefore, the FSA-based models significantly outperformed the Lasso and BOR models on these datasets. We can also confirm that the Lasso and BOR models showed very similar results for regression estimates. When evaluating the results of each model, the FSA ANN models were among the top-performing groups for six of the 10 datasets, and the baseline ANN models were in the top five of the 10 datasets.

*5.3 Forecasting return movements*

The FSA-based LR model was not applied in this case, as it was basically an FSA model with a logistic loss function. Accuracy and AUC metrics were used for model classification comparisons, and the experimental results are presented in Tables 11–12. A star represents the best model for each column, and those that are not significantly worse than the best model ($p > 0.05$) are shown in bold.

Table 11: Return movement forecasting. Accuracy values for the 10 datasets evaluated

| Model/Data | TTH | ETH | BTC | RPL | TSL | NSDQ | NKK | NYSE | S&P500 | GLD |
|---|---|---|---|---|---|---|---|---|---|---|
| LR | **0,524** | 0,515 | 0,507 | 0,517 | **0,515** | **0,743** | 0,515 | **0,539** | 0,518 | 0,503 |
| ANN | 0,487 | 0,498 | 0,485 | **0,527** | 0,494 | **0,745** | 0,535 | 0,516 | **0,538** | 0,496 |
| CNN | 0,502 | 0,517 | 0,490 | **0,535*** | 0,513 | **0,740** | 0,527 | 0,526 | 0,526 | 0,495 |
| LSTM | 0,482 | 0,529 | **0,523*** | 0,528 | 0,490 | 0,523 | 0,522 | 0,498 | 0,525 | 0,530 |
| XGB | 0.477 | 0.489 | 0.489 | 0.502 | **0.530*** | **0.758** | 0.462 | 0.473 | 0.459 | 0.509 |
| FSA | 0.501 | 0.512 | **0.523** | **0.532** | 0.507 | **0.748** | 0.519 | 0.499 | 0.530 | **0.538*** |
| LAS | 0.463 | 0.491 | 0.461 | 0.461 | 0.443 | 0.734 | 0.420 | 0.441 | 0.456 | 0.453 |
| FSA-ANN | 0,486 | **0,535** | **0,523*** | **0,535*** | **0,513** | 0,734 | 0,528 | 0,526 | 0,525 | 0,528 |
| FSA-CNN | 0,490 | 0,521 | **0,515** | **0,527** | **0,508** | 0,747 | **0,553*** | 0,529 | 0,525 | 0,528 |
| FSA-LSTM | **0,527*** | **0,558*** | 0,509 | 0,515 | 0,463 | 0,521 | 0,528 | **0,534** | **0,554*** | 0,502 |
| FSA-XGB | 0.416 | **0.544** | **0.514** | 0.495 | 0.496 | **0.759*** | 0.456 | 0.483 | **0.537** | 0.453 |
| BOR-LR | 0.493 | 0.464 | **0.513** | **0.533** | 0.461 | 0.530 | 0.442 | 0.512 | 0.477 | 0.492 |
| BOR-ANN | 0.484 | 0.467 | **0.513** | 0.522 | 0.488 | 0.509 | 0.544 | 0.512 | 0.468 | 0.500 |
| BOR-CNN | 0.468 | 0.473 | **0.516** | 0.506 | 0.470 | 0.530 | 0.540 | 0.521 | 0.483 | 0.503 |
| BOR-LSTM | 0.470 | 0.513 | 0.495 | 0.503 | **0.505** | 0.532 | 0.453 | 0.454 | 0.521 | 0.479 |
| BOR-XGB | 0.441 | 0.519 | 0.503 | 0.506 | **0.500** | 0.536 | 0.485 | 0.487 | 0.471 | 0.474 |
| LAS-LR | 0.466 | 0.481 | 0.476 | 0.509 | 0.451 | 0.696 | 0.481 | 0.526 | 0.483 | 0.484 |
| LAS-ANN | 0.495 | 0.497 | 0.489 | 0.513 | 0.458 | 0.722 | 0.495 | **0.564*** | 0.504 | 0.478 |
| LAS-CNN | 0.496 | 0.478 | **0.514** | 0.473 | 0.446 | **0.739** | 0.461 | 0.512 | 0.519 | 0.496 |
| LAS-LSTM | 0.464 | 0.519 | 0.471 | **0.526** | **0.517** | 0.539 | 0.446 | 0.476 | 0.512 | 0.481 |
| LAS-XG | **0.520** | 0.449 | 0.463 | 0.484 | **0.500** | 0.734 | 0.452 | 0.515 | 0.498 | 0.480 |

In terms of accuracy, Table 11 shows that the FSA models significantly outperformed all non-FSA models in three of the 10 datasets (i.e., ETH, NKK, and GLD). Furthermore, they outperformed all Lasso and BOR models on four datasets (i.e., ETH, NKK, SP500, and GLD), meaning that, on these datasets, the FSA model made significant contributions. In addition to these results, linear FSA models were within the top-performing group on five of the 10 datasets (i.e., BTC, RPL, TSL, NSDQ, and GLD). The linear FSA model outperformed all other estimated models on the GLD dataset. The FSA-based models, including linear FSA, were in the top-performing group for all datasets. Moreover, at least one of the Lasso-based models was in the top-performing group on six of the 10 datasets; however, they did not make a significant contribution to any of the six datasets. The Lasso-based ANN model outperformed all other models on the NYSE dataset, excluding the LR and FSA LSTM models. Similarly, the BOR models were in the top-performing group on only three datasets as they did not bring significant contributions, as among the plain models in the LR, ANN, CNN, and LSTM sets, one was in the top-performing group. Many models on the NSDQ dataset obtained the highest accuracies. The FSA-XGB model provided the best accuracy of 75.9%, and LR, ANN, CNN, XGB, and linear FSA were in the top-performing accuracies. Therefore, the FSA did not provide a significant contribution at this point.

In terms of AUC, Table 12 shows that FSA-based models were in the top-performing group for nine of the 10 datasets, as they significantly outperformed the plain models on the NKK dataset. The Lasso-based models were in the top-performing group for three of the 10 datasets (i.e., TSL, NSDQ, and

NYSE). However, there was no significant contribution based on the group in which it was located. The BOR models were in the top-performing group on two of the 10 datasets. These models did not provide significant contributions, as they were in the same group as the plain model. For the NSDQ dataset, the BOR models lagged very far behind the plain, FSA, and Lasso-based models in terms of AUC and accuracy.

Table 12: Return movement forecasting. AUC values for the 10 datasets evaluated.

| Model/Data | TTH | ETH | BTC | RPL | TSL | NSDQ | NKK | NYSE | S&P500 | GLD |
|---|---|---|---|---|---|---|---|---|---|---|
| LR | **0,533** | 0,516 | 0,510 | 0,515 | 0,514 | **0,823** | 0,521 | **0,547** | 0,531 | 0,505 |
| ANN | 0,500 | 0,547 | 0,494 | **0,566*** | 0,499 | **0,826*** | 0,556 | 0,539 | **0,563** | **0,540** |
| CNN | 0,527 | 0,512 | 0,477 | 0,524 | **0,525** | 0,821 | 0,547 | 0,492 | 0,530 | 0,501 |
| LSTM | 0,495 | **0,560** | **0,536*** | 0,536 | 0,511 | 0,502 | 0,549 | 0,526 | 0,474 | **0,538** |
| XGB | 0.475 | 0.481 | 0.492 | 0.494 | **0.527** | 0.758 | 0.447 | 0.503 | 0.461 | 0.510 |
| FSA | 0,487 | 0,533 | 0,524 | **0,533** | 0,507 | **0,818** | 0,529 | 0,534 | 0,528 | **0,542*** |
| LAS | 0.453 | 0.501 | 0.469 | 0.469 | 0.493 | 0.798 | 0.443 | 0.432 | 0.456 | 0.474 |
| FSA-ANN | 0,496 | 0,532 | 0,521 | **0,547** | 0,512 | **0,810** | 0,538 | 0,536 | 0,552 | **0,538** |
| FSA-CNN | 0,517 | 0,516 | **0,531** | 0,543 | 0,509 | **0,811** | **0,573*** | 0,527 | 0,532 | 0,531 |
| FSA-LSTM | **0,542*** | **0,569*** | 0,503 | 0,509 | **0,544*** | 0,528 | 0,538 | 0,523 | **0,576*** | 0,522 |
| FSA_XG | 0.423 | 0.530 | 0.509 | 0.482 | 0.494 | 0.759 | 0.436 | 0.481 | 0.531 | 0.454 |
| BOR-LR | 0.481 | 0.451 | 0.518 | 0.518 | 0.464 | 0.516 | 0.526 | 0.507 | 0.499 | 0.483 |
| BOR-ANN | 0.475 | 0.437 | 0.521 | 0.528 | 0.424 | 0.465 | **0.565** | **0.547** | 0.504 | 0.512 |
| BOR-CNN | 0.465 | 0.440 | 0.513 | 0.514 | 0.453 | 0.508 | 0.553 | 0.536 | 0.493 | 0.511 |
| BOR-LSTM | 0.493 | 0.520 | 0.510 | 0.498 | 0.457 | 0.497 | 0.509 | 0.507 | 0.475 | 0.481 |
| BOR-XG | 0.452 | 0.510 | 0.501 | 0.519 | 0.504 | 0.524 | 0.499 | 0.503 | 0.462 | 0.486 |
| LAS-LR | 0.473 | 0.490 | 0.469 | 0.501 | 0.487 | 0.718 | 0.476 | **0.541** | 0.497 | 0.462 |
| LAS-ANN | 0.496 | 0.526 | 0.499 | 0.501 | 0.463 | 0.799 | 0.522 | **0.557*** | 0.500 | 0.498 |
| LAS-CNN | 0.497 | 0.487 | 0.486 | 0.466 | 0.461 | **0.811** | 0.487 | 0.525 | 0.542 | 0.497 |
| LAS-LSTM | 0.500 | 0.500 | 0.496 | 0.506 | **0.541** | 0.548 | 0.440 | 0.491 | 0.479 | 0.502 |
| LAS-XG | 0.524 | 0.453 | 0.464 | 0.472 | 0.497 | 0.730 | 0.469 | 0.526 | 0.499 | 0.475 |

When evaluating the results of each model, linear FSA and FSA-based CNN models were in the top-performing group on five of the 10 datasets. The results showed that an accuracy level between 51 and 76% was attained for all 10 datasets. A slight improvement in forecasting performance can lead to increased profitability; therefore, obtaining a realistic prediction is very important for investors. There are similar studies that make comparisons to this study. However, it is not possible to make a precise comparison due to our unique input–output mapping process, our applied algorithm, and the applied input feature sets. Kara et al. (2011) reported 71%–82% accuracy levels using 10 common technical indicators. He and Fan (2021) reported 82%–87%, and Patel et al. (2015b) reported 65%–91% accuracy levels using different input feature sets. Altman (1968) used a similar input–output mapping procedure and obtained a 48% accuracy level in the third period using conventional statistical models. Our proposed model applied forecasting returns and return movements using three periods of lagged variables to obtain realistic results. It is believed that this study will increase researchers' attention to these areas based on the benefits of our new feature-selection algorithm for financial forecasting. This study is conducted as a whole process. Data obtaining, constructing a new input space by calculating technical indicators and lagged variables, preprocessing, feature selection, and financial forecasting. While evaluating this complete process, it is easy to say our proposed models and research methodology provided successful estimations. However, there are a few limitations that may have significant effects on the process, such as applying online learning algorithms. There was no restriction on data access thanks to the financial data availability and open public policy. Our expected results completely matched the obtained.

*6 Discussions*

The cryptocurrency market has recently received a great deal of attention due to its offerings of alternative investment assets versus the stock market. The technology used for cryptocurrency blockchain systems is key to its attraction to global investors. The significant chaotic behavior of the cryptocurrency market, as well as modern stocks, creates a great deal of risk for investors. Hence, they require highly professional and detailed recommendation systems to assist them with their portfolio management.

This work investigated the use of ML feature-selection techniques for improving forecasting performance and introduced the first FSA application for this purpose, which combined ANN, CNN, LSTM, and XGBoost models to obtain linear and nonlinear solutions for feature selection. These feature-selection models were studied based on their classification performance, with the aim of forecasting the up and down movement of stocks. Via regression, the aim was to forecast actual stock returns. The main research question of this study is to investigate the effect of the FSA algorithm on ML financial forecasting models. The combined proposed models significantly improved the forecasting performance of ML algorithms. This means that the proposed models greatly contributed to finance and ML literature. By following this procedure, it is possible to construct financial recommendation systems to help financial investors with their investment decisions. Online learning algorithms for ML models may also significantly increase the model performance.

A total of 10 financial time-series datasets were constructed for our experimental validation, in which different technical indicators and lags were used to obtain a 1,105-dimensional feature vector. Technical indicators with specific periods and their first five lags were used to construct the input feature pool for forecasting. As response variables were investigated, 1-, 2-, and 3-day-ahead returns and movements were examined (Yıldırım et al., 2021). Because the 1$^{st}$ and 2$^{nd}$ days response variables were predicted with high accuracies and high $R^2$ values, the 3$^{rd}$-day-ahead trend and its logarithmic return values were chosen as response variables. Specific datasets were used based on the price value of the selected cryptocurrencies and stocks, as these parameters are the most important for investors and analysts.

In summary, this study provides an estimation of financial market forecasting. Estimated models were used to forecast the financial market using existing information based on technical indicators. As a result, we obtained reliable forecasts, even if they lacked high accuracy. The financial time-series problem is complicated, and reliable models are difficult to construct; hence, the accuracy level of estimations can vary greatly. Considering our results, we have shown that the EMH theory is valid for the financial market in the ML age. However, we still need comprehensive statistical tests and analyses to prove our validity.

*7 Conclusions*

This paper investigated whether the FSA algorithm can improve the forecasting performance of certain ML techniques based on a selection of 10 key financial time-series financial datasets. The Lasso and BOR feature-selection algorithms, which are popular for financial data analysis, were evaluated and compared to our FSA implementation to provide a broad set of results.

The findings indicate that the FSA algorithm improved or at least did not reduce forecasting performance. Moreover, the FSA method provided better estimations than the Lasso and BOR models in six of the 10 datasets applied to regression tasks. Moreover, the FSA models outperformed the plain ML models on three of the 10 datasets. The best FSA-based LSTM models greatly reduced MSE values by more than 20% on the TTH dataset. The FSA-based ANN and CNN models provided approximately 10% improvement in terms of MSE for the SP500 and GLD datasets. The BOR and Lasso models provided similar estimation results regarding MSE values for three of the 10 datasets, placing them in the top-performing group. The FSA-based ANN models provided the best estimations of MSE in six of the 10 datasets. Moreover, the linear FSA model was placed in the top-performing group on five of the 10 datasets. These results support our linear model usage justification for feature selection, as presented in the methodology section.

In addition to regression estimation, plain ML models provided the best accuracy values for three of the 10 datasets (i.e., BTC, RPL, and TSL). The FSA-based models were in the top-performing group on nine of the 10 datasets, with higher accuracy values between 52 and 76%. Linear FSA provided the best accuracy value for the GLD dataset at 54% by outperforming all other estimated models. The FSA-based CNN model for the NKK dataset, the FSA-based LSTM, and the ANN models for the ETH dataset made significant contributions due to their significant procedural advantages. The BOR models were in the top-performing group on three of the 10 datasets. The Lasso models were in the top-performing group on six of the 10 datasets, providing the highest accuracies for the NYSE dataset. The FSA-based LSTM and Lasso ANN models provided great improvements by increasing the accuracy values by approximately 10%. In seven of the 10 datasets, the FSA-based models provided more than 7% improvements in accuracy. The FSA-based models provided the best accuracies between 53 and 76% on eight of the 10 datasets, and the Lasso-based ANN provided the best accuracy values of 56% for the NYSE dataset. The linear FSA model was in the top-performing group on five out of ten datasets, with the best accuracy of 75%. However, the FSA-based XGB model gave the best accuracy of 76% for the NSDQ dataset.

The Lasso and BOR provided higher MSEs on seven datasets and lower accuracy values on four datasets. Therefore, we can conclude that the FSA model is superior to the Lasso and BOR, especially for financial regression tasks. One of the most important conclusions drawn from this study is that the prediction performance of the BOR algorithm remains at its minimum level. In many cases, this algorithm is reduced in terms of its forecasting performance with ML models. Reducing the dimensionality of the input feature space is needed to clearly understand the financial problem and is useful for reducing computation times and memory requirements. Using the same input feature space may not be helpful in increasing the prediction performance of different datasets. This is why the feature-selection process of our method provides significant contributions to analysts and investors.

There is plenty of room for advancements in this study, such as employing and evaluating traditional statistical models. Moreover, a multiclass classifier should be used to provide recommendations for different risk levels and other useful comments compared with binary classifiers. Furthermore, feature-generation processes, which have received very little attention in the literature, may be helpful in improving the predictive ability of algorithms. Other ML algorithms, such as autoencoders, attention models, and NLP methods, may increase forecasting accuracy while providing more powerful results by considering the context and semantics of financial analyst recommendations and investor reactions. Owing to the sensitivity of the financial market to politics and policies, NLP techniques provide a unique opportunity to glean significant insights into causal factors. Deep learning and ensemble learning methods may also be used to implement superior estimation models, albeit at significant computational costs. Moreover, governance considerations (e.g., rule of law, transparency, and accountability) should be examined as input features to better understand how they contribute to the problem.

When constructing a real-time investor recommendation system that uses online learning algorithms and expert systems, the experts themselves should be involved. Researchers are always working on algorithms to increase model performance. However, ground-truth knowledge must be incorporated for training and verification purposes. The number of stock and cryptocurrency data points increases every second and entails many chaotic movements. Hence, an ultracomprehensive data preprocessing system is needed to help researchers and investors better understand the behaviors of market forces.

**List of abbreviations**

- ANN    Artificial neural network
- ARIMA Auto regressive integrated moving average
- AUC    Area under curve
- BOR    Boruta feature selection
- CNN    Convolutional neural network
- FSA    Feature selection with annealing
- GBM   Gradient boosting machine

| | |
|---|---|
| GRNN | Generalized regression neural network |
| LDA | Linear discriminant analysis |
| LR | Logistic regression |
| LSTM | Long short-term memory |
| MAE | Mean absolute error |
| MAPE | Mean absolute percentage error |
| ML | Machine learning |
| MLP | Multi-layer perceptron |
| MSE | Mean squared error |
| NARX | Nonlinear autoregressive with exogenous variable |
| NB | Naïve bayes |
| ReLU | Rectified linear unit |
| RF | Random forest |
| RNN | Recurrent neural network |
| SVM | Support vector machine |
| SVR | Support vector regression |
| TA-Lib | Technical analysis library |
| XGB | Extreme gradient boosting |

Trend Prediction. In: 2017 International Conference on Tools with Artificial Intelligence Time-weighted. pp 1210–1217